# Improving Generalization Ability of Genetic Programming: Comparative Study

Tejashvi R. Naik
Information Technology Department,
Dharmsinh Desai University, Nadiad, INDIA
tejashvi_naik1989@yahoo.com

Vipul K.Dabhi
Information Technology Department,
Dharmsinh Desai University,Nadiad, INDIA
vipul.k.dabhi@gmail.com

**Abstract:**
**In the field of empirical modeling using Genetic Programming (GP), it is important to evolve solution with good generalization ability. Generalization ability of GP solutions get affected by two important issues: bloat and over-fitting. Bloat is uncontrolled growth of code without any gain in fitness and important issue in GP. We surveyed and classified existing literature related to different techniques used by GP research community to deal with the issue of bloat. Moreover, the classifications of different bloat control approaches and measures for bloat are discussed. Next, we tested four bloat control methods: Tarpeian, double tournament, lexicographic parsimony pressure with direct bucketing and ratio bucketing on six different problems and identified where each bloat control method performs well on per problem basis. Based on the analysis of each method, we combined two methods: double tournament (selection method) and Tarpeian method (works before evaluation) to avoid bloated solutions and compared with the results obtained from individual performance of double tournament method. It was found that the results were improved with this combination of two methods.**

*Index Terms--* Genetic Programming (GP), Symbolic Regression, Multi-valued Regression, Generalization, Bloat, Empirical Modeling, Evolutionary computation (EC), bloat.

## 1. INTRODUCTION

Like many arbitrary-sized representations in evolutionary computation (EC), a genetic programming (GP) individual tends to grow in size when no code growth measures are being applied to it. The growth is relatively independent of significant increase in fitness. The use of arbitrary –length representation in EC Presents a special challenge to search process in evolution.The phenomenon of bloat is the uncontrolled growth of individuals in population. The problem with bloat is that: it slows down the search process, hampers breeding and consumes memory. Bloat occurs almost in all evolutionary computation methods like neural networks, automata, genetic programming, Genetic algorithm (GA) and rule sets. For any Machine Learning (ML) technique, to generalize means, the technique is expected to generate a solution that could achieve same generalization performance on unseen data as obtained on training data. Since, there is a lack of exact theoretical explanation of bloat, the genetic programming has relied on different approaches, each having its own advantages and disadvantages. The most common technique to deal with the problem of bloat is to reject children's during breeding whose depth exceeds than some maximum tree depth by koza (in 1992) and second is parsimony pressure methods[1]. The Minimum Description Length (MDL) [5] approach to improve generalization ability of solutions induced by GP suggests promoting evolution of simpler solutions compare to complex solutions [2]. The approach suggests that it is more likely that complex solutions may contain specific information from training data and thus may over-fit it compared to simpler solutions. The paper begins by reviewing the issue of bloat. Next, we have reviewed why bloat is important and theories of bloat. Next, section 3 describes various bloat control methods that we have considered for comparison and experimentation. Next, section is experimental setup in section 4 and then lastly results and conclusion.

## 2. BLOAT IN GP

The phenomenon of uncontrolled growth in size of individual without any significant improvement in fitness is known as bloat. Early theory of bloat concentrated on the existence of introns, areas of code that can be removed without altering the fitness value of the solution [3]. From a theoretical point of view there are several theories that are either based on intron or non-intron theory of bloat: *the hitchhiking theory, the defense against crossover theory, the replication accuracy theory, the removal bias theory are based on introns and the nature of program search space theory, fitness causes bloat , modification depth point are non-intron theory [3, 6].*

The hitchhiking theory [3], states that random selection in combination with random crossover does not cause code growth and therefore it is concluded that fitness is the cause of the increasing of the size. Defense against crossover theories [3], go a step further. They argue that the role played by the intron is that of increasing the number of nodes of the tree, making it more difficult to destroy with crossover. The replication accuracy theory [3], states that the success of a solution lies on its ability to have offspring that are identical to the parent fitness-wise. The removal bias theory [3] states that, given that redundant data tends to be low in the tree (i.e. closer to the leaves than to the root) and applying crossover to redundant data does not modify the fitness of a solution, the evolution will favor the replacement of small branches. To maintain fitness, the removed sub tree must be contained within the inviable region. Since the inserted sub-tree can have any size, offspring are bigger than average whileretaining the fitness of their parents.

The nature of program search space theory [3], is the only theory not based on introns. The idea here is that above a certain size, the fitness does not vary with size. Since in the search space there are more big tree structures than small ones during the search process the GP will tend to find bigger trees.Fitness causes bloat [6]: there are many more longer ways than shorter ways to represent the same program, so a natural drift occurs to longer programs. Modification point depth [6]: When a genetic operator modifies an individual, the deeper the modification point the smaller the change in fitness. Small changes are less likely to be disruptive, so there is preference for deeper modification points, and consequently a preference for larger trees (removal bias). Crossover bias theory [3], explains bloat by assuming that crossover operator on its own does not produce growth or shrinkage in size of solutions. Repeated application of crossover operations push the population towards a particular distribution of tree sizes, where small size trees have high frequency than longer ones. Since small size trees are not useful in solving problem, larger size trees have a selective advantage. Thus, average solution size of population increases.

## 3. BLOAT CONTROL METHODS:

We have considered four bloat control methods for comparative study and experimental purpose based on the analysis of paper [3]. The different measures of bloat are: mean fitness, mean tree size, execution time, by taking difference of structural complexities of evolved solution and target solution[2], can also be measured based on relationship between average model length growth and average fitness improvement at current generation compared to respective values at generation zero in [2]. Our bloat control measures were: mean fitness and mean tree size. The reason for considering these two measures is, as the population grows it consumes more memory and creates difficulties and the final individual returned at the end would be excessively large. The four bloat control methods are discussed in this section. The bloat control methods are classified into: (1) parametric parsimony pressure methods, ranked based methods and non-parametric methods, (2) Direct/Indirect methods [3], (3) Adaptive/Non-Adaptive[3]. Unlike other techniques we discussed above parsimony pressure is not GP specific and can be used with any arbitrary sized representation. Parametric parsimony is a linear combination of size and fitness. In ranked based bloat control methods, the minor difference in fitness values matters as the rank of the individual is considered not the actual fitness for calculation. The non-parametric are based on simple modification of tournament selection to consider both size and fitness, but not together as a combined parametric equation [7].The issue with parametric parsimony method is they are hard to tune to prevent size from dominating fitness late in evolutionary process or to compensate for problem dependent non-linearity in raw fitness function [7]. Direct method controls bloat by evaluating special genetic operators and indirect method controls bloat by accepting/rejecting the solutions modified by genetic operators or through selection. Depending on whether the parsimony co-efficient values are fixed or vary during the GP run, the methods are classified into adaptive/non-adaptive.

### 3.1    Tarpeian Method:

In this method, before the evaluation process some individuals with above average size are assigned bad fitness with some probability W i.e. Kill-proportion and therefore reduces the number of evaluations. Tarpeian is a parametric method.

### 3.2    Lexicographic Parsimony Pressure with Direct Bucketing:

In this method, the number of buckets b are defined beforehand and then each individual is assigned a rank from 1 to b. First the population p is sorted based on fitness and then bottom the [total individuals/b] individuals are placed in worst bucket then second bucket is filled up until there are no individuals left in population.

### 3.3    Lexicographic Parsimony Pressure with Ratio Bucketing:

In this method, we need to define a bucket ratio [1/r] beforehand. This method is a slight improvement over direct bucketing in that , ratio bucketing guarantees that low fitness individuals will pushed in to large buckets and place high fitness individuals in smaller buckets. We put 1/R worst individuals into lower buckets along with the remaining individuals with same fitness as the best individual in the bucket. We then put next ¼ next worst individuals in the next bucket and so on until there are no individuals left in the population. The fitness of each individual is set to the rank assigned to the bucket holding it.

### 3.4 Double Tournament Method:

This method is two layer hierarchy:.
1. Qualifying tournament.
2. Final tournament.

In the qualifying tournament the individuals are selected based on fitness (if do-fitness-first=true) or vice a versa. The winners of this tournament now become contestants for final tournament based on length of individual or vice a versa. The selection is parameterized by: fitness tournament size(size 2), parsimony tournament size(size) and do-fitness-first which indicates weather fitness tournaments are used in qualifier tournament or final tournament. Double tournament is a non-parametric method.

## 4. METHODOLOGY

The toolkit used for testing bloat control methods was ECJ –evolutionary computation framework written in java. We used ECJ toolkit as it is a open source, platform independent, extensible, multi-objective support, merresine twister random number generator and emphasis of this toolkit was more towards genetic programming (GP). In ECJ, parameter file is the main part of the system in which all the details necessary for the GP problem to execute are defined. There are four main parameter files: ec.params, Simple.params, koza.params and finally the parameter file of our application. For testing of bloat control methods we need to use parsimony package of ECJ and then make changes in koza.params for the parsimony methods to work. The hierarchy of the parameter file is shown in **figure 1.** For Tarpeian method we need to make changes in parameter file of our problem. To add more than one variable in a problem or to add a new function, we have to make java file for that variable or function and add this variable or function in parameter file of the problem. We have added four terminals {x1, x2, x3, x4} for five dimensional parameter problem and two function files Tan.java and sqrt.java for regression, sextic, five dimensional parameter problems. For multi-valued regression problems we have added seven function files for the functions F={/, log, exp, sqrt, tan, sin, cos}. In ECJ, random inputs are generated for every run within a specified range, to make it static we need to define inputs in the problem file and write seed.0=time in ec.params file

## 5. EXPERIMENTAL SETUP

We have considered four bloat control methods: Tarpeian methods, lexicographic parsimony pressure with direct bucketing, lexicographic parsimony pressure with ratio bucketing, double tournament. Problems on which these bloat control methods were applied are: (1) symbolic regression problem which used the function, $x^4 + x^3 + x^2 + x$, with no ephemeral random constant. Symbolic Regression asks tree to fit a real valued function within a domain [-10,10]. Terminal set used is {x}. Since, Symbolic Regression suffers from a very large amount of inevitable code, so many individuals in the population have identical fitness. (2) Symbolic Regression problem with constants used a function $3.0x^3 + 11.0x^2 + 14.0x + 6.0$ in domain of [-10,10]. Terminal set used for this problem was {x}. (3) Sextic problem used a function $x^6 - 2.0x^4 + x^2$ in range of [-10, 10]. Terminal set used for this problem was {x} (4) Multi-valued valued regression problem used a function $x^2y + xy + y$ in range of [-10,10]. Terminal set used for this problem was {x,y}. (5) Multi-valued Regression problem with constants used a function $3x^2y + 4xy + y$ in range of [-10, 10]. Terminal set used for this problem was {x,y}. (6) Function finding on five dimensional parameter space which used a function (Math.sin(x)* Math.cos(x1))/ Math.sqrt(Math.exp(x2)) + Math.tan(x3-x4) in range of [-1,1]. Terminal set used for this problem was {x, x1, x2, x3, x4}. Regression problem with constants used a function $3x^2y + 4xy + y$ in range of All experiments used a population size of 1000 and number of generations was 50. Each method was experimented for 40 runs. The fitness function used is 1.0f/(1.0f+fitness) And mean tree size is the size of the best individual of this generation. Then we analyzed the results of these techniques and combined two techniques based on performance in avoiding bloated solutions /reducing the bloat. The common parameter settings for all methods are shown in table 1. For lexicographic direct bucketing method, selection method used is Bucket tournament selection and number of buckets B=10, 25, 50, 100, 250. For ration bucketing method, selection method is Ratio Bucket tournament selection with bucket ratio R= 2, 3, 4, 5, 6, 7, 8, 9, 10. For double tournament method, selection method is double tournament selection with parsimony tournament size D= 1.0, 1.1, 1.2, 1.3, 1.4, 1.5, 1.6, 1.7, 1.8, 1.9, 2.0. For Tarpeian method we need to specify Tarpeian statistics file in parameter file of our application and also define the kill-proportion=0.0, 0.1, 0.2, 0.3, 0.4, 0.5.

**Table 1 common GP parameter settings for all problems**

| Parameter | Value |
|---|---|
| Tournament size | 7 |
| Crossover /mutation probability | 0.8/0.1 |
| Population size | 1000 |
| Max depth | 17 |
| functions | {+,-,/,*,tan, sin, cos, log, exp , sqrt} |
| No. of generations | 50 |

**Figure 1 hierarchy of parameter files used in ECJ**

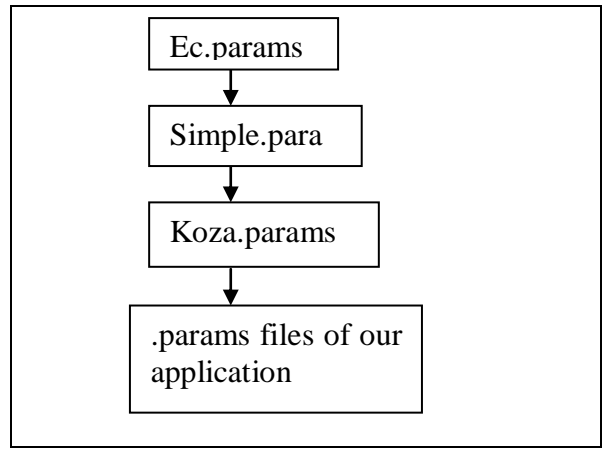

**Table 2 classification of bloat control approaches**

| Characteristics | Code-editing | Size/depth limit | Tarpien | Lexicographic parsimony pressure | Lexicographic parsimony pressure with ratio bucketing | Lexicographic parsimony pressure with Direct bucketing | Double tournament |
|---|---|---|---|---|---|---|---|
| Direct/Indirect | Direct | Indirect | Indirect | Indirect | Indirect | Indirect | Indirect |
| Parametric/rank-based/non-parametric | Non-parametric | parametric | parametric | Non-parametric | Ranked based | Rank based | Non-parametric |
| Adaptive/non-adaptive | Non-adaptive | Non-adaptive | Non-adaptive | Non-adaptive | Non-adaptive | Non-adaptive | Non-adaptive |
| Phase of GP | Other | Breeding | Evaluation | Selection | Selection | Selection | Selection |

# 6. EXPERIMENTAL RESULTS:

This section presents the results obtained in comparative study of bloat control methods. The average mean fitness and mean tree size for all methods are shown in following tables and graphs. The results of direct bucketing and tarpeian method are shown in **table 3** and results of ratio bucketing and double tournament are shown in **table 4.** The functioning of the method is considered superior if the mean tree size of the population is reduced while maintaining the fitness. In Tarpeian method, the kill-proportion with 0.4, 0.5 gave significant results for all problems. In Ratio bucketing, almost all values uniformly performed well. Consistent values for ratio bucket were r= 6, 7, 9. Direct bucketing did not gave significantly good results for all methods. Consistent values for direct bucketing were number of buckets= 250 and 500. Double tournament performed well for all problems with consistent value of parsimony size D=1.6, 1.8, 1.9. The results obtained for regression problem without constant of all methods are shown in **figure 2** and **figure 4**.The results obtained for regression problem with constant of all methods are shown in **figure 3** and **figure 5.** The results obtained for sextic problem are shown in **figure 6** and **figure 8** and results obtained for multi-valued regression problem without constant of all methods are shown in figure **figure 7** and **figure 9**. The results obtained for multi-valued regression problem with constant are shown in **figure 10** and **figure 12** and the results obtained for five dimensional parameter problem are shown in **figure 11** and **figure 13.**

**Table 15 Results obtained with tarpeian method and double tournament method**

| Problems | Parsimony co-efficient (num-of-buckets)-B | Lexicographic parsimony pressure with Direct Bucketing | | Tarpeian method | | Kill-proportion |
|---|---|---|---|---|---|---|
| | | Mean fitness | Mean tree size | Mean fitness | Mean tree size | |
| 1) Regression problem without constant | B=25 | 0.141101 | 17.74728 | 0.137067 | 16.43746 | W=0.1 |
| | B=50 | 0.160197 | 18.42969 | 0.136458 | 16.12046 | W=0.3 |
| | B=100 | 0.151319 | 19.29648 | 0.146161 | 14.36611 | W=0.4 |
| | B=250 | **0.140487** | **17.43793** | **0.120954** | **13.72386** | W=0.5 |
| | B=500 | 0.148719 | 18.50298 | | | |
| 2) Regression problem with constant | B=10 | **0.011909** | 90.2935 | 0.009781 | 91.895 | W=0.0 |
| | B=250 | 0.016069 | **88.2185** | 0.010609 | 92.2215 | W=0.3 |
| | B=500 | 0.012054 | 89.1115 | **0.009419** | **88.082** | W=0.4 |
| 3) Sextic problem | B=10 | 0.086025 | 38.80267 | 0.037521 | 27.73994 | W=0.4 |
| | B=100 | 0.086867 | **35.95256** | **0.021589** | **25.34243** | W=0.5 |
| | B=250 | **0.080547** | 39.79703 | | | |
| | B=500 | 0.084895 | 40.75359 | | | |
| 4) Multi-valued Regression problem without constant | B=10 | **0.145219** | **15.14128** | 0.16934 | 17.30057 | W=0.0 |
| | B=50 | 0.187668 | 16.77978 | 0.148886 | 18.05082 | W=0.1 |
| | B=100 | 0.165034 | 15.37124 | 0.151441 | 19.24608 | W=0.2 |
| | B=250 | 0.157513 | 16.98063 | **0.136272** | **17.01919** | W=0.3 |
| | B=500 | 0.147851 | 17.19274 | 0.131799 | 19.39159 | W=0.5 |
| 5) Multi-valued – regression problem with constant | B=10 | 0.082656 | 40.03821 | 0.137067 | 16.43746 | W=0.1 |
| | B=25 | 0.078852 | 40.42887 | 0.151269 | 16.84731 | W=0.2 |
| | B=100 | 0.070531 | 36.86088 | 0.136458 | 16.12046 | W=0.3 |
| | B=250 | **0.067426** | **36.74538** | 0.146161 | 14.36611 | W=0.4 |
| | B=500 | 0.080948 | 38.41678 | **0.120954** | **13.72386** | W=0.5 |
| 6) Five dimensional parameter problem | B=50 | **0.04854** | **33.032** | 0.046877 | 16.5435 | W=0.4 |
| | B=100 | 0.048785 | 34.83 | **0.046572** | **13.7025** | W=0.5 |
| | B=250 | 0.048721 | 35.412 | | | |
| | B=500 | 0.048707 | 33.551 | | | |

**Table 3 Results obtained with Ratio bucketing and Double tournament method on six problems**

| Problems | Parsimony co-efficient(Bucket Ratio )-1/r | Lexicographic Parsimony Pressure with Ratio Bucketing | | Double tournament method | | Parsimony co-efficient (size d) |
|---|---|---|---|---|---|---|
| | | Mean fitness | Mean tree size | Mean fitness | Mean tree size | |
| 1) Regression problem without constant | R=2 | 0.146646 | 19.91979 | 0.14104 | 18.5613 | D=1.3 |
| | R=3 | 0.139141 | 18.15887 | 0.143004 | 17.53308 | D=1.4 |
| | R=4 | 0.136726 | 19.53164 | 0.149582 | 18.61901 | D=1.5 |
| | R=5 | 0.148686 | 17.82583 | 0.151539 | 18.65905 | D=1.6 |
| | R=6 | **0.133762** | 19.19882 | 0.149953 | **17.12107** | D=1.8 |
| | R=7 | 0.139882 | 18.35347 | 0.141389 | 17.26386 | D=1.9 |
| | R=9 | 0.150355 | **17.07202** | 0.151197 | 17.62298 | D=2.0 |
| | R=10 | 0.145201 | 17.72344 | **0.140192** | 17.79118 | d=4.0 |
| 2) Regression problem with constant | R=5 | 0.011803 | 90.132 | 0.014552 | **85.431** | D=1.3 |
| | R=6 | 0.010827 | 91.7455 | 0.013707 | 86.597 | D=1.5 |
| | R=7 | **0.002353** | **83.3735** | **0.01119** | 88.3455 | D=1.6 |
| | R=9 | **0.014001** | 89.3365 | 0.011045 | 87.0425 | D=1.8 |
| | | | | 0.012242 | 88.602 | D=1.9 |
| | | | | 0.011412 | 86.957 | D=2.0 |
| 3) Sextic problem | R=4 | 0.098189 | 37.8191 | 0.083186 | 36.90242 | d=1.1 |
| | R=5 | 0.089054 | 41.02677 | **0.083039** | **36.31977** | d=1.3 |
| | R=6 | 0.089264 | 36.67724 | 0.091278 | 38.59487 | D=1.5 |
| | R=9 | 0.081446 | **34.57259** | 0.085078 | 37.36005 | D=1.6 |
| | R=10 | 0.10854 | 37.97659 | 0.088776 | 39.3942 | D=1.8 |
| 4) Multi-valued regression problem without constant | R=3 | **0.144779** | 20.17105 | 0.170345 | 13.69177 | D=1.4 |
| | R=4 | 0.161199 | **14.72529** | **0.148797** | 14.97966 | D=1.6 |
| | R=5 | 0.146252 | 15.75461 | 0.16481 | 15.68272 | D=1.7 |
| | R=6 | 0.168938 | 20.50862 | 0.160857 | **13.1206** | D=1.8 |
| | R=9 | 0.166478 | 16.1898 | 0.164346 | 14.38245 | D=1.9 |
| 5) Multi-valued regression problem with constant | R=2 | 0.063713 | 38.45679 | **0.090064** | 37.41417 | D=1.5 |
| | R=3 | 0.110917 | 39.33158 | 0.093295 | **35.89481** | D=1.6 |
| | R=5 | 0.079341 | **37.77995** | 0.096223 | 34.36675 | D=1.8 |
| | R=6 | 0.078159 | 39.82679 | 0.094126 | 37.52701 | D=1.9 |
| | R=7 | 0.081479 | 37.82622 | 0.097830 | 36.88492 | D=2.0 |
| | R=8 | **0.059206** | 38.34232 | | | |
| | R=9 | 0.0675 | 37.86902 | | | |
| 6) Five dimensional parameter problem | R=4 | 0.04889 | 37.376 | 0.048494 | 31.6445 | D=1.6 |
| | R=6 | 0.048853 | 36.49385 | **0.048141** | **29.0315** | D=1.7 |
| | R=7 | 0.04862 | 39.178 | 0.048135 | 30.3425 | D=1.8 |
| | R=8 | 0.048675 | 37.905 | 0.048223 | 30.516 | D=1.9 |
| | R=9 | **0.048285** | **33.8475** | | | |
| | R=10 | 0.048889 | 37.53 | | | |

**Figure 2** mean fitness obtained for different bloat control methods for regression problem without constant

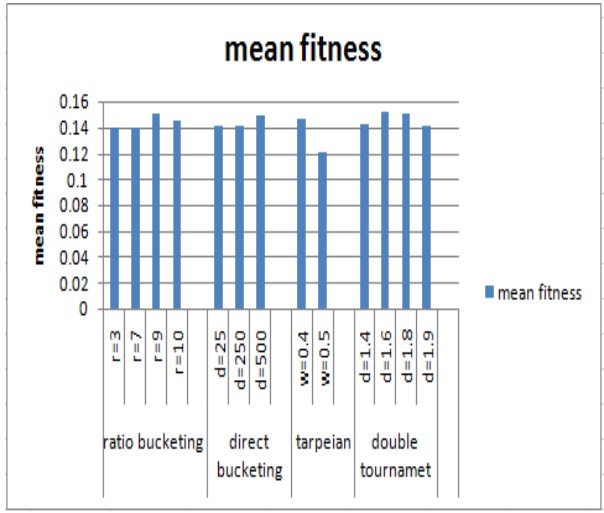

**Figure 3** mean fitness obtained for different bloat control methods for regression problem with constant

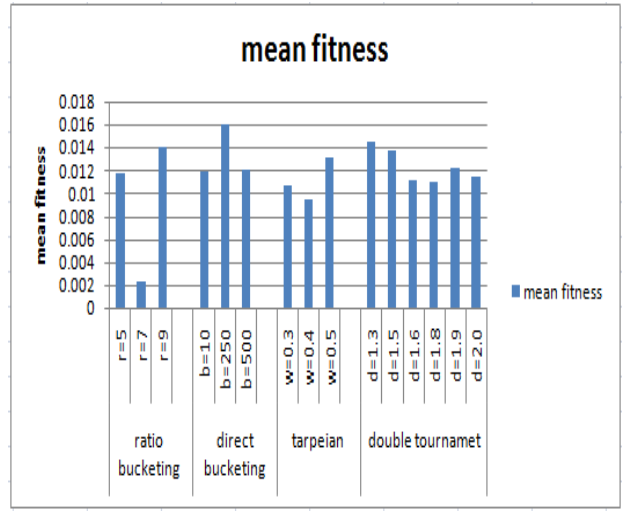

**Figure 4** mean tree size obtained for different bloat control methods for regression without constant

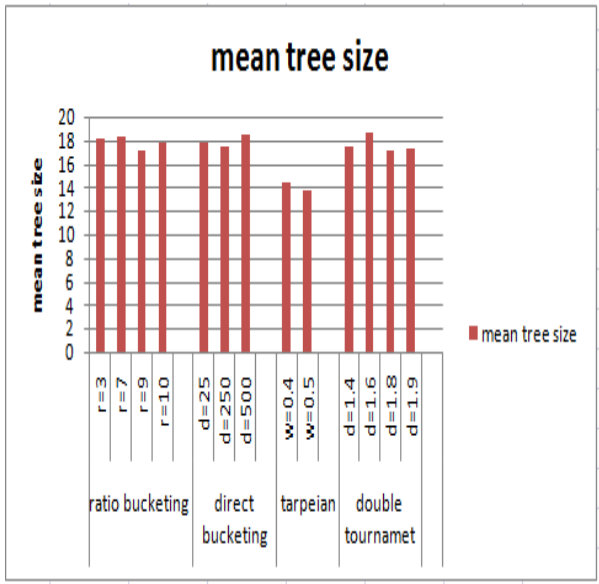

**Figure 5** mean tree size obtained for different bloat control methods for regression problem with constant

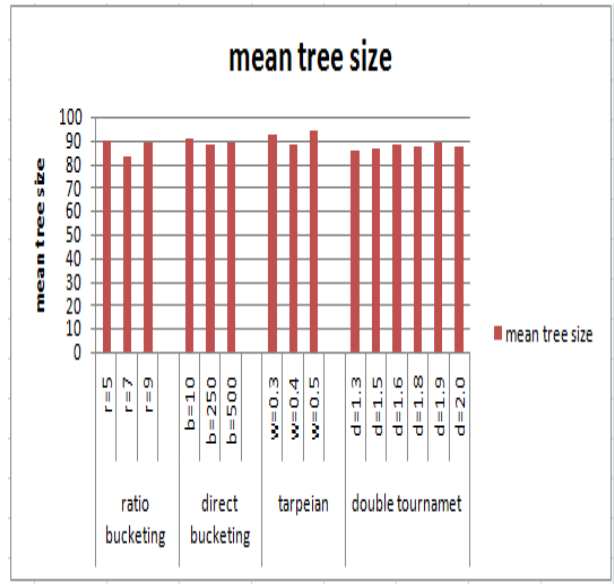

Figure 6 mean fitness obtained for different bloat Control methods for sextic problem

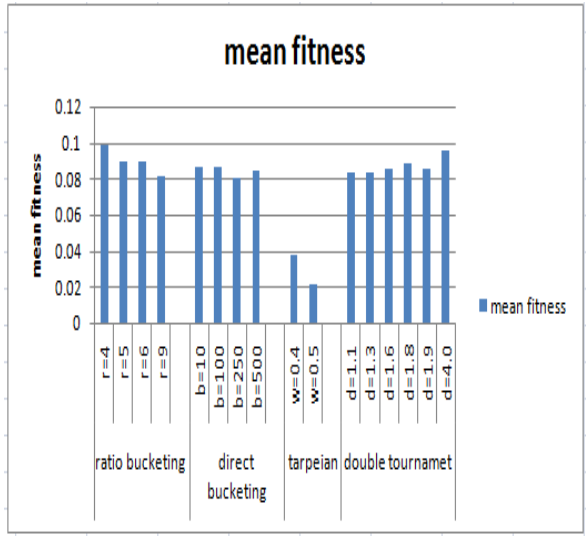

figure 7 mean fitness obtained for different bloat control methods for multi-valued regression problem without constant

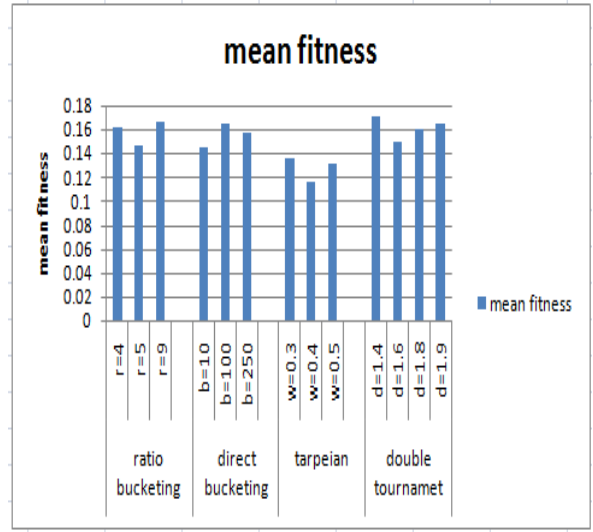

Figure 8 mean tree size obtained for different bloat Control methods for sextic problem

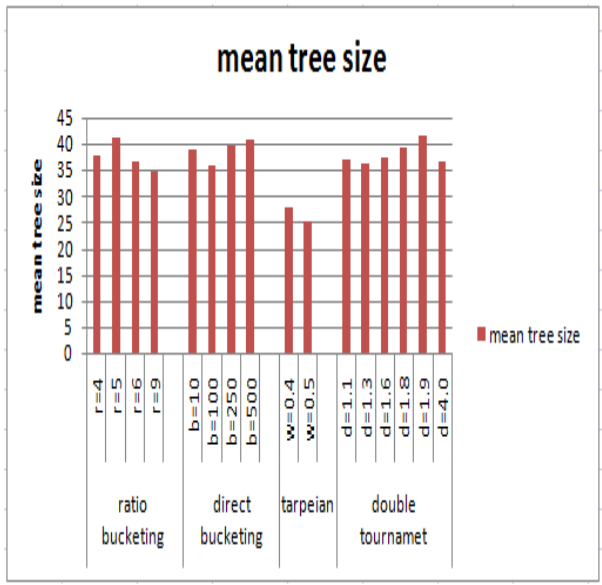

Figure 9 mean tree size obtained for different bloat control methods for multi-valued regression problem without constant

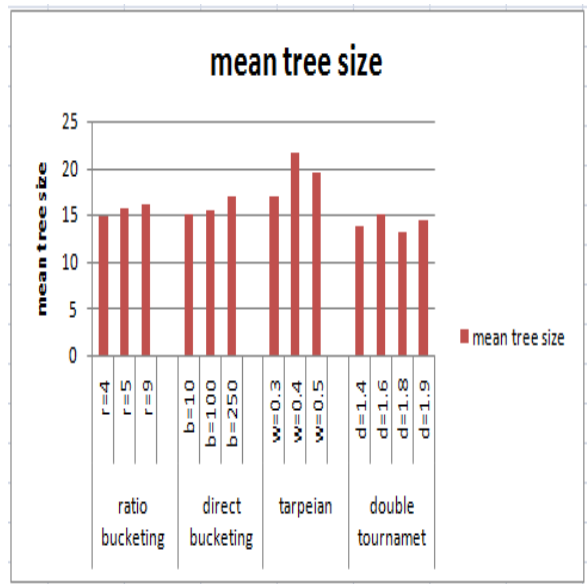

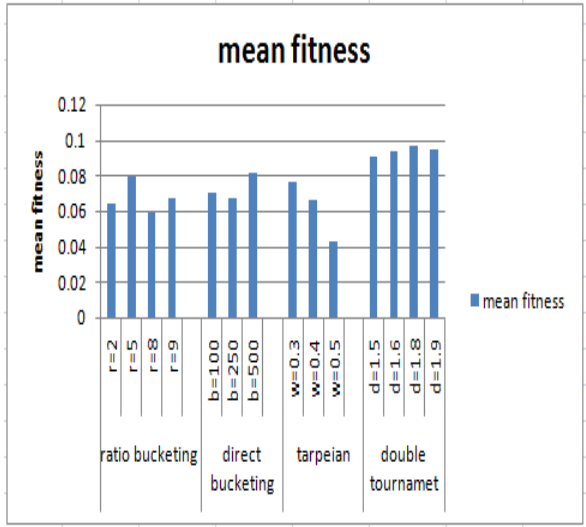

Figure 10 mean fitness obtained for different bloat control methods for multi-valued regression Problem with constant

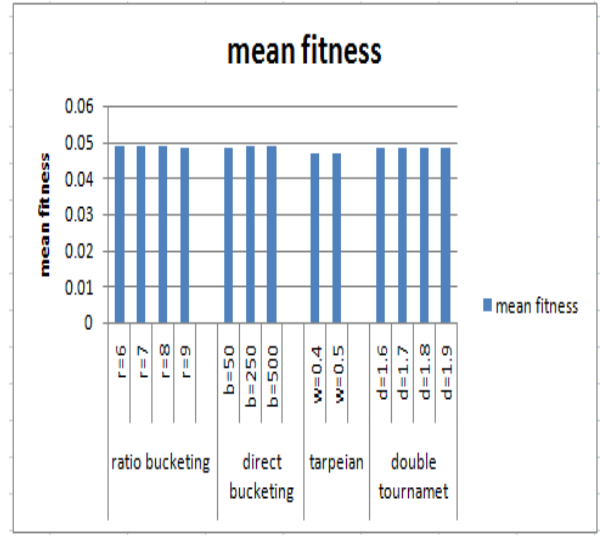

figure 11 mean fitness obtained for different bloat control methods for five- dimensional parameter problem

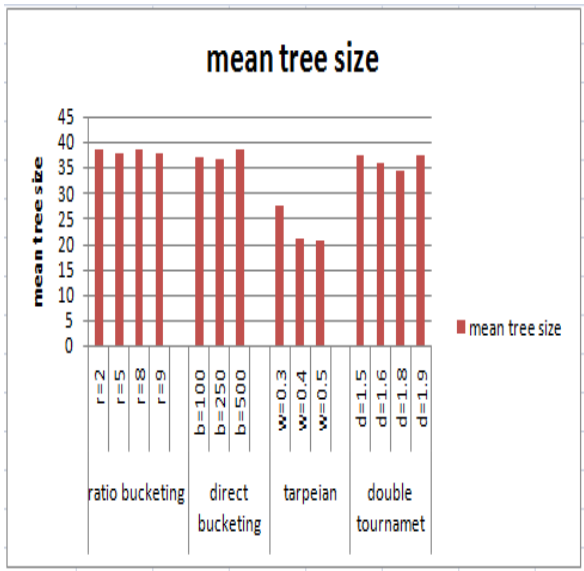

Figure 12 mean tree size obtained for different bloat control Methods for multi-valued regression problem with constant

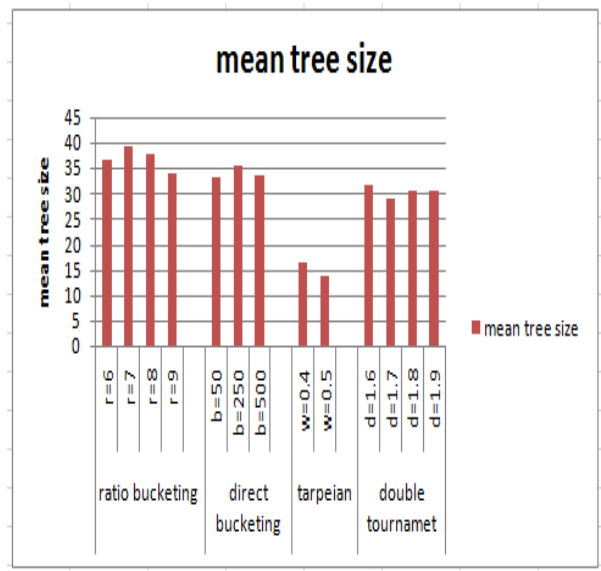

Figure 13 mean tree size for different bloat control methods for five-dimensional parameter problem

**NOTE**: We performed experiments of four bloat control methods: Tarpeian method, double tournament, direct bucketing and ration bucketing on six different problems. The results are analyzed based on two bloat measures: mean fitness and mean tree size. For Regression problem without constant, all methods performed well. For regression problem with constant, ratio bucketing and double tournament outperformed Tarpeian and direct bucketing. For sextic problem, Tarpeian method outperformed all other methods. For multi-valued regression problem, ratio bucketing and double tournament outperformed other methods. For multi-valued regression with constant, double tournament and Tarpeian performed well, ratio bucketing method and direct bucketing were not able to reduce boat. For five dimensional parameter problem, double tournament and Tarpeian outperformed other methods. Based on the results of performance of bloat control methods on per-problem bases, it was found that double tournament and Tarpeian performed well on all the problems.

## 7. DOUBLE TOURNAMENT AUGMENTED WITH TARPEIAN METHOD

We combined double tournament method with Tarpeian method based on the performance of individual methods. Double tournament works as a selection method and Tarpeian is added as statics file in parameter file of problem as it operated before evaluation. We selected values of the size in double tournament and kill-proportion in tarpeain based on the results obtained from individual performance of the methods. The parameter setting for the combination of double tournament and tarpeian methods are shown in **table 5**

**Table 5 GP parameter setting for double tournament method combined with tarpeian**

| Parameter | Value |
| --- | --- |
| Selection method | Double tournament |
| Parsimony co-efficient d, kill-proportion | D=1.6,1.8,1.9 with w=0.3,0.4,0.5 |
| Tournament size (f) | 7 |
| Do-fitness-first | True |
| Crossover /mutation probability | 0.8/0.1 |
| Population size | 1000 |
| Max depth | 17 |
| functions | {+,-,/,*,tan, sin, cos, log, exp, sqrt } |

**Table 6 Results obtained with double tournament combined with tarpeian method on six problem**

| Applications | Size | Double tournament | | Kill-prportion | Tarpein method | | Settings | Double tournament + Tarpeian | |
|---|---|---|---|---|---|---|---|---|---|
| | | Mean-fitness | Mean tree size | | Mean-fitness | Mean tree size | | Mean-fitness | Mean tree size |
| 1. Regression without constant | D=1.4 | 0.143004 | 17.53308 | W=0.3 | 0.136458 | 16.1204 | D=1.6, W=0.3 | 0.14802 | 13.4588 |
| | D=1.6 | 0.151539 | 18.65905 | W=0.4 | 0.146161 | 14.3661 | D=1.8, W=0.4 | 0.11673 | 12.9918 |
| | D=1.8 | 0.149953 | **17.12107** | W=0.5 | | | D=1.9, W=0.5 | | |
| | D=1.9 | **0.141389** | 17.26386 | | **0.120954** | **13.7238** | | **0.09604** | **12.7977** |
| 2. Regression with constant | D=1.3 | 0.014552 | **85.431** | W=0.3 | 0.010609 | 92.2215 | D=1.6, W=0.3 | 0.00751 | 69.144 |
| | D=1.6 | **0.008147** | 88.3455 | W=0.4 | **0.009419** | **88.082** | D=1.8, W=0.4 | 0.00683 | 54.121 |
| | D=1.8 | 0.013707 | 87.0425 | W=0.5 | | | D=1.9, W=0.5 | | |
| | D=1.9 | 0.01119 | 88.602 | | **0.013148** | 94.1465 | | **0.00359** | **44.8725** |
| 3. Sextic | D=1.3 | **0.083039** | **36.31977** | W=0.3 | 0.061589 | 34.6859 | D=1.6, W=0.3 | 0.02537 | 33.8265 |
| | D=1.6 | 0.098086 | 37.36005 | W=0.4 | 0.037521 | 27.7399 | D=1.8, W=0.4 | 0.02352 | 25.5446 |
| | D=1.8 | 0.091278 | 38.59487 | W=0.5 | | | D=1.9, W=0.5 | | |
| | D=1.9 | 0.085078 | 41.54281 | | **0.021589** | **25.3424** | | **0.00574** | **16.2773** |
| 4. Multi-valued-regression without constant | D=1.4 | 0.170345 | 13.69177 | W=0.3 | 0.136272 | **17.0191** | D=1.6, W=0.3 | 0.15890 | 19.5833 |
| | D=1.6 | **0.142234** | 14.97966 | W=0.4 | **0.115466** | 21.62104 | D=1.8, W=0.4 | 0.14062 | **15.0026** |
| | D=1.8 | 0.148797 | **13.1206** | W=0.5 | | | D=1.9, W=0.5 | | |
| | D=1.9 | 0.16481 | 14.38245 | | 0.131799 | 19.39159 | | **0.09327** | 18.5023 |
| 5. Multi-valued regression with constant | D=1.6 | **0.0932954** | 35.89481 | W=0.3 | 0.136458 | 16.12046 | D=1.6, W=0.3 | 0.08278 | 27.4201 |
| | D=1.8 | 0.0962239 | **34.36675** | W=0.4 | 0.146161 | 14.36611 | D=1.8, W=0.4 | 0.08364 | 23.2872 |
| | D=1.9 | 0.094126 | 37.52701 | W=0.5 | | | D=1.9, W=0.5 | | |
| | D=2.0 | 0.097830 | 36.88492 | | **0.120954** | **13.7238** | | **0.04531** | **19.6713** |
| 6. Five dimensional parameter space | D=1.7 | 0.048141 | **29.0315** | W=0.3 | 0.047506 | 20.78 | D=1.6, W=0.3 | 0.0471 | 20.2665 |
| | D=1.6 | 0.048494 | 30.3425 | W=0.4 | 0.046877 | 16.5435 | D=1.8, W=0.4 | 0.04669 | 13.8235 |
| | D=1.8 | **0.048135** | 30.3425 | W=0.5 | | | D=1.9, W=0.5 | | |
| | D=1.9 | 0.048223 | 30.516 | | **0.046572** | **13.7025** | | **0.04630** | **9.967** |

**Figure 14 mean fitness of double tournament (method 1) method compared with double tournament combined with Tarpeian (method2)**

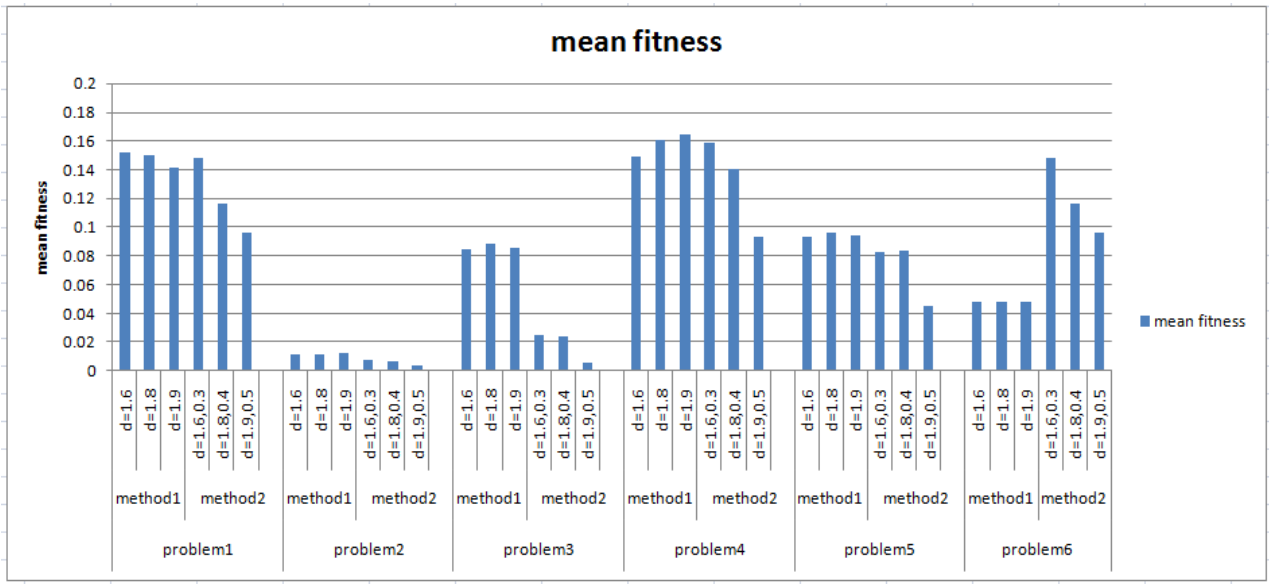

**Figure 15 mean tree size of double tournament (method 1) compared with double tournament combined with tarpeain method (method 2)**

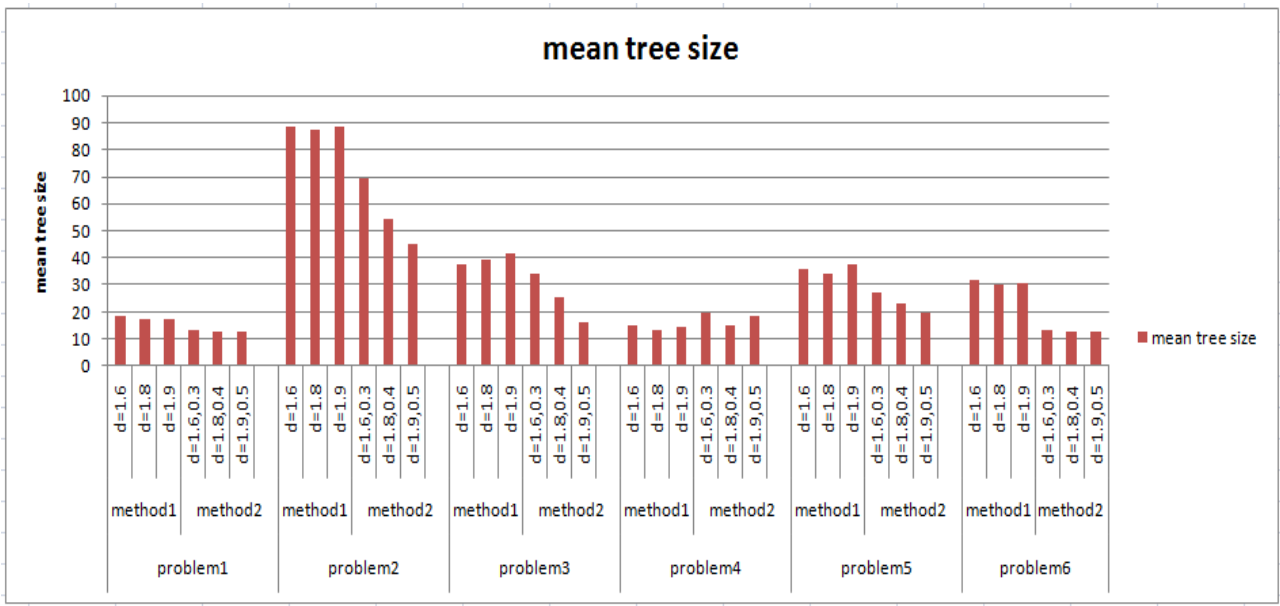

## 8. CONCLUSION

In this paper, we discussed the problem of bloat and importance of bloat in genetic programming. Then we analyzed the different bloat control theories based on introns and parsimony pressure methods. We performed experiments with four bloat control methods: Tarpeian method, double tournament, direct bucketing and ratio bucketing on six different problems. The performance of the methods are examined based on two measures of bloat i.e. fitness and mean tree size. For Regression problem without constant, tarpeian method outperformed other methods. For regression problem with constant, ratio bucketing and double tournament outperformed Tarpeian and direct bucketing. For sextic problem, Tarpeian method outperformed all other methods. For multi-valued regression problem without constant, ratio bucketing and double tournament outperformed other methods. For multi-valued regression with constant, double tournament and Tarpeian performed well. For five dimensional parameter problem, double tournament and Tarpeian outperformed other methods. From the results, it was found that there was very minor difference between the fitness values, but this minor difference also matters as the bloat control methods were ranked based and non-parametric except Tarpeian method. In tarpeian method, the kill-proportion value w=0.4, 0.5 performed consistently well on all the problems and for kill-proportion(W) value higher than 0.5 results in reduced tree size but with worst fitness, therefore we have not considered those values for comparison. Based on the results obtained from different bloat control methods on per-problem bases, it was found that double tournament and Tarpeian performed well on all the problems. From the analysis of the results obtained, we combined double tournament and Tarpeian and tested it on all six problems. The results of combination of these two methods are presented and compared with individual performance of double tournament method. It was found that the combination of these two methods outperformed the performance of individual methods except on multi-valued regression problem without constant.